\documentclass[runningheads]{llncs}

\usepackage{cite}
\usepackage[pdftex]{graphicx}
\usepackage[export]{adjustbox}

\usepackage{hyperref}

\usepackage{lineno}

\usepackage{url}
\usepackage{booktabs}
\usepackage{graphicx}
\usepackage{subfigure}
\usepackage{amsmath}
\usepackage{amssymb}
\usepackage{dblfloatfix}
\usepackage{color}
\usepackage{tabularx}
\usepackage{soul}

\newcommand{\blue}[1]{{\color{blue} #1}}
\newcommand{\anon}[1]{\blue{anonymous}}
\renewcommand{\anon}[1]{{#1}}
\DeclareMathOperator{\softmax}{softmax}

\DeclareMathOperator{\softplus}{softplus}
\DeclareMathOperator{\LSTM}{LSTM}
\newcommand{\mybold}[1]{\boldsymbol{\mathbf{#1}}}

\newcommand{\bb}{\mybold{b}}
\newcommand{\cb}{\mybold{c}}

\newcommand{\hb}{\mybold{h}}

\newcommand{\vb}{\mybold{v}}
\newcommand{\xb}{\mybold{x}}

\newcommand{\mub}{\mybold{\mu}}

\newcommand{\Hb}{\mybold{H}}

\newcommand{\Xb}{\mybold{X}}
\newcommand{\Yb}{\mybold{Y}}
\newcommand{\Wb}{\mybold{W}}

\newcommand{\pib}{\mybold{\pi}}

\begin{document}

\title{Automated Surgical Activity Recognition with One Labeled Sequence}


\author{Robert DiPietro and Gregory D. Hager}
\institute{Department of Computer Science, Johns Hopkins University, Baltimore, MD, USA}

\maketitle

\begin{abstract}
Prior work has demonstrated the feasibility of automated activity recognition in robot-assisted surgery from motion data. However, these efforts have assumed the availability of a large number of densely-annotated sequences, which must be provided manually by experts. This process is tedious, expensive, and error-prone. In this paper, we present the first analysis under the assumption of scarce annotations, where as little as \emph{one annotated sequence} is available for training. We demonstrate feasibility of automated recognition in this challenging setting, and we show that learning representations in an unsupervised fashion, before the recognition phase, leads to significant gains in performance. In addition, our paper poses a new challenge to the community: how much further can we push performance in this important yet relatively unexplored regime?

\end{abstract}

\section{Introduction}
\label{introduction}

The advent of robot-assisted surgery has spawned many new research areas, in large part because it allows the capture of high-quality surgical-motion data at scale. Two examples are objective performance assessment \cite{reiley2008,vedula2016} and automated feedback for trainees \cite{chen2016}, which have the potential to transform surgical training curricula and in turn improve patient outcomes \cite{birkmeyer2013}. An important prerequisite task toward these goals and others is \emph{surgical activity recognition}, where we aim to automatically segment and label surgical-motion data according to the activities being performed by a surgeon or trainee.

Significant progress has been made in surgical activity recognition, especially within the context of simulated training \cite{ahmidi2017,dipietro2016,dipietro2019}, an important part of current training curricula \cite{ase2001}. Though promising, these approaches have relied on large amounts of annotated data, which, unlike the surgical-motion data itself, must be provided \emph{manually} by experts. This process is expensive, error-prone, and often subjective, especially when carried out at scale.

\begin{figure}[t]
\centering
\includegraphics[scale=0.4]{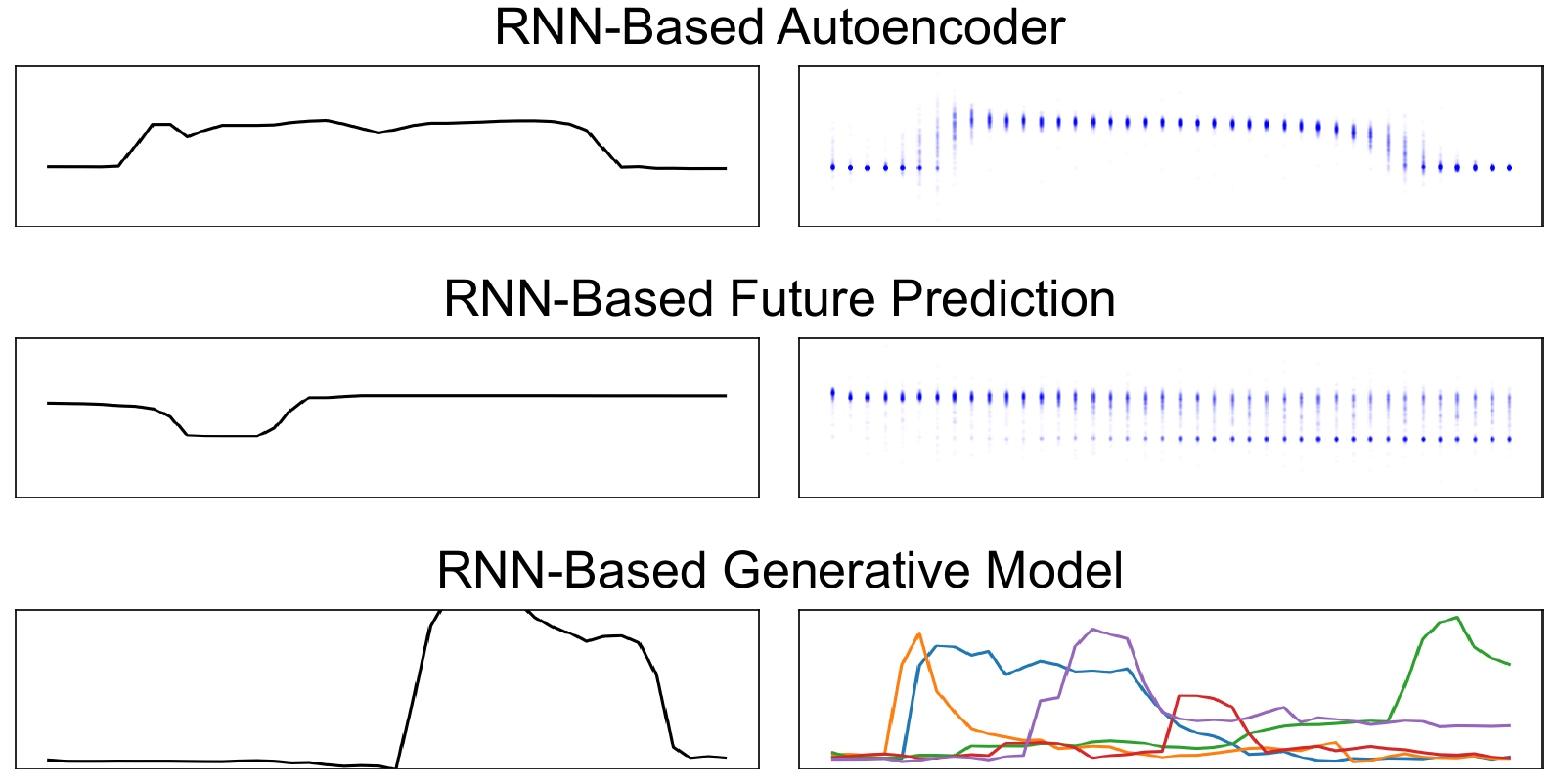}
\caption{\textbf{Example predictions for the three tasks considered for unsupervised representation learning.} On the left, we see the input to each model; and on the right, we see sampled predictions. The autoencoder reconstructs the input window; the future-prediction model predicts a window assuming independence over time steps, conditioned on a previous window; and the generative model predicts coherent futures of any length, conditioned on the entire past (5 sampled trajectories are shown).}
\label{samples}
\end{figure}

Despite these difficulties, literature has largely ignored activity recognition in the context of scarce annotations. To our knowledge, the only exceptions have been in the form of preprints, and have focused on video-based recognition rather than motion-based recognition \cite{bodenstedt2017,yengera2018,yu2018}. The most closely related work to this paper is \cite{dipietro2018}, in which we explored unsupervised learning of surgical motion via future prediction. Two limitations of \cite{dipietro2018} are 1. considering applications to motion-based search rather than the predominant task of activity recognition and 2. considering models that assume independence over time, which lead to blurry, incoherent future prediction (see Figure \ref{samples}).

\begin{figure}[t]
\centering
\includegraphics[scale=0.45]{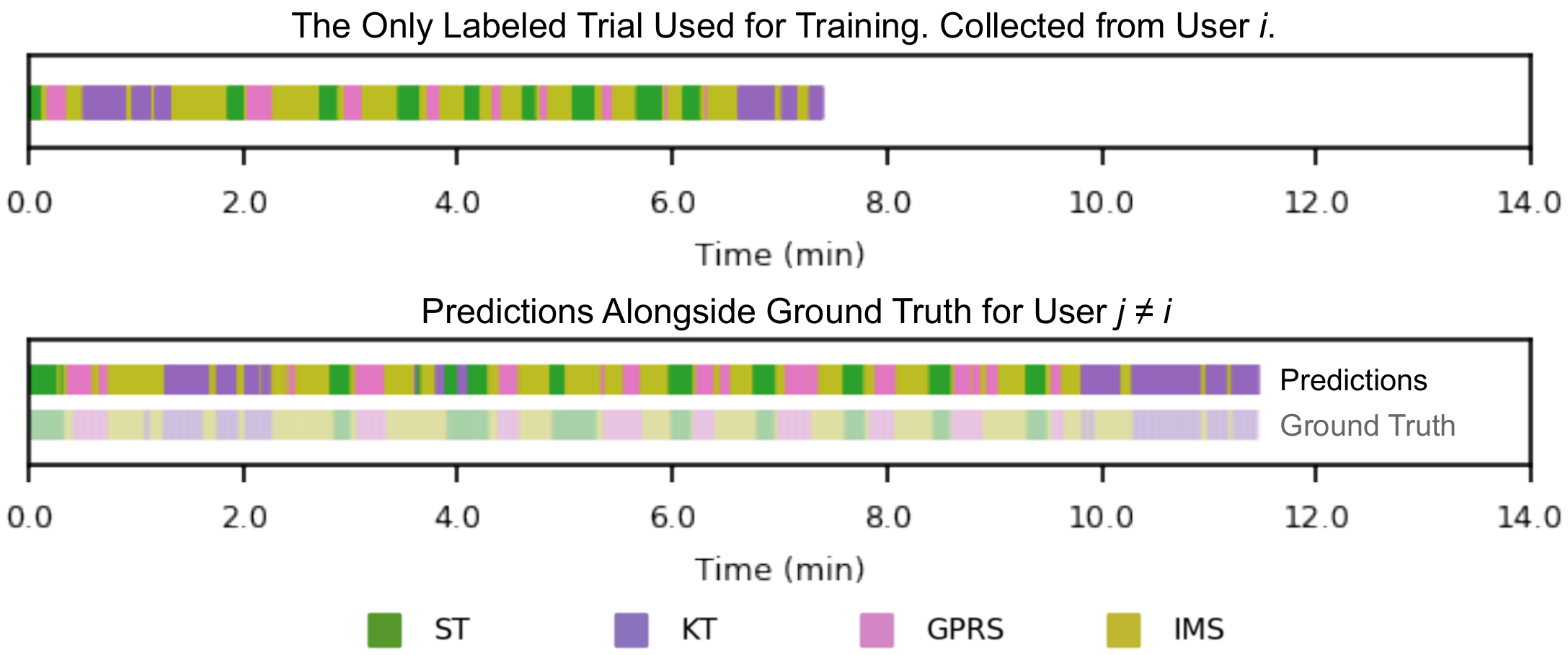}
\caption{\textbf{Example predictions for maneuver recognition, using only a single labeled sequence for training.} Here representations were learned with the RNN-Based Future Prediction model prior to recognition. It exhibits a representative error rate (19.4\%) and an edit distance that is worse than average (40.7\%). Results are similar for the RNN-Based Generative Model. The activities are \emph{suture throw} (ST), \emph{knot tying} (KT), \emph{grasp pull run suture} (GPRS), and \emph{intermaneuver segment} (IMS).}
\label{teaser}
\end{figure}

The primary contributions of this paper are 1. demonstrating the feasibility of activity recognition in annotation-scarce settings (e.g., Figure \ref{teaser}); 2. showing that preceding recognition with unsupervised representation learning leads to significant gains in recognition performance for both maneuver recognition and gesture recognition (e.g., Figures \ref{misticerrorbarplot} and \ref{jigsawserrorbarplot}); 3. introducing a probabilistic generative model over surgical motion that makes no simplifying assumption of independence over time (Section \ref{methods}); and 4. showing that this generative model is especially important when used to recognize activities that exhibit fine-grained structure over short time scales (Figure \ref{jigsawserrorbarplot}).

\section{Methods}
\label{methods}

We let $\Xb \equiv \{\xb_t\}_{1}^{T}$ denote a sequence of kinematics, with each $\xb_t \in \mathbb{R}^{n_x}$ (containing, e.g., joint velocities), and we let $\Yb \equiv \{y_t\}_{1}^{T}$ denote a corresponding sequence of activities, where each $y_t$ is an integer (specifying an activity). We aim to learn a mapping from $\Xb$ to $\Yb$, which corresponds to joint segmentation and classification. This is accomplished in two phases. The first is representation learning, where we learn a transformation from $\Xb$ to $\tilde{\Xb} \equiv \{\tilde{\xb}_t\}_{1}^{T}$ through an auxiliary task that requires no annotations. The second is recognition, in which we learn a mapping from $\tilde{\Xb}$ to $\Yb$.

\paragraph{Representation Learning Overview.} We consider three models for representation learning. The first two mimic those from \cite{dipietro2018}, where they were used for motion-based search. The first, the RNN-Based Autoencoder, aims to reconstruct a given window of kinematics through a bottleneck; and the second, RNN-Based Future Prediction, models a \emph{window} of future motion conditioned on a \emph{window} of previous motion, \emph{assuming conditional independence over time}. The third model, which we introduce here and refer to as the RNN-Based Generative Model, is a generative model over all of $\Xb$, \emph{without} any simplifying assumption of independence over time. Here we describe the RNN-Based Generative Model in detail, and we refer the reader to \cite{dipietro2018} for a full description of the RNN-Based Future Prediction and RNN-Based Autoencoder models.

\paragraph{The RNN-Based Generative Model.} The RNN-Based Generative Model models the full joint distribution over $\Xb$ by making use of the chain rule,
\begin{equation}
\label{replearningdist}
p(\xb_1, \xb_2, \ldots, \xb_T) = p(\xb_1) p(\xb_2 \mid \xb_1) p(\xb_3 \mid \xb_1, \xb_2) \cdots p(\xb_T \mid \xb_1, \ldots, \xb_{T-1})
\end{equation}
This representation facilitates the use of recurrent neural networks and mixture density networks \cite{bishop1994} to model each factor in the product sequentially: At time $t$, we use long short-term memory (LSTM) \cite{hochreiter1997,gers2000fg} to map from the previous kinematics vector $\xb_{t-1}$ and previous hidden state $\hb_{t-1}$ to a new hidden state (the LSTM cell $\cb_t$ is omitted here for simplicity):
\begin{equation}
\hb_t = \LSTM(\xb_{t-1}, \hb_{t-1})
\end{equation}
Next, we map from $\hb_t$ to the parameters that govern the distribution $p(\xb_t \mid \xb_1, \ldots, \xb_{t-1})$. This can be any reasonable distribution of our choosing. Following \cite{dipietro2018} for simplicity and fair comparisons, we map from $\hb_t$ to the parameters of a mixture of Gaussians with diagonal covariance via
\begin{align}
\pib_t &= \softmax(\Wb_\pi \, \hb_t + \bb_\pi) \\
\mub_t^{(c)} &= \Wb_\mu^{(c)} \, \hb_t + \bb_\mu^{(c)} \\
\vb_t^{(c)} &= \softplus(\Wb_v^{(c)} \, \hb_t + \bb_v^{(c)})
\end{align}
with the conditional distribution over $\xb_t$ then specified as
\begin{equation}
p(\xb_t \mid \xb_1, \ldots, \xb_{t-1}) =
\sum_c \pi_t^{(c)} \, \mathcal{N}\!\left(\xb_t \,;~ \mub_t^{(c)}, \vb_t^{(c)}\right)
\end{equation}
All weight matrices $\Wb$ and all biases $\bb$, in both the LSTM and in the mapping from hidden states to distribution parameters, are learned by maximizing (the logarithm of) Eq. \ref{replearningdist}. After training, the sequence of hidden states from the LSTM, $\Hb$, is used as inputs in the recognition phase: $\tilde{\Xb} = \Hb$.

\paragraph{Recognition Overview.} We discriminatively model $p(\Yb \mid \tilde{\Xb})$ using the multilayered, bidirectional LSTM architecture from \cite{dipietro2019}. First, $\tilde{\Xb}$ is transformed into a sequence of hidden states $\Hb$ (simply LSTM hidden states, distinct from $\Hb$ of the previous section), and these hidden states are then mapped in the standard fashion (through affine transformations) to the parameters that govern the categorical distributions over each $p(y_t \mid \tilde{\Xb})$. Training is carried out by maximizing conditional likelihood under this model, or equivalently by minimizing cross entropy. Please see \cite{dipietro2019} for more detail.

\section{Experiments}
\label{experiments}

\begin{figure}[t]
\centering
\includegraphics[width=3.5in]{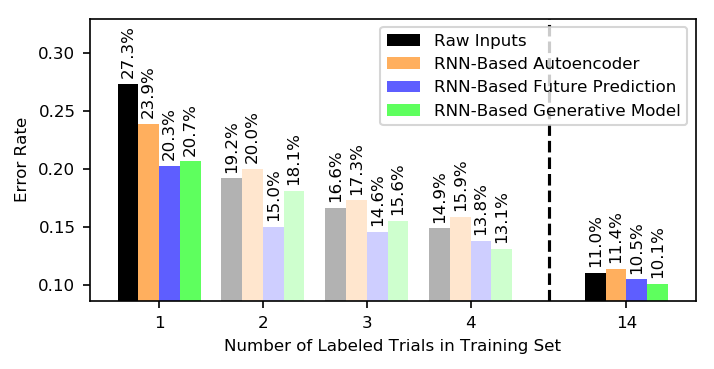}
\caption{\textbf{MISTIC-SL Maneuver Recognition: Error rate vs. number of labeled trials.} The bottom of the $y$ axis is set to 8.7\%, the best published result using LSTM ($\sim$36 labeled trials). The non-transparent results are reported over exhaustive, deterministic splits (see Section \ref{experimentaldesign}) which can be compared to in future work.}
\label{misticerrorbarplot}
\end{figure}

\begin{figure}[t]
\centering
\includegraphics[width=3.5in]{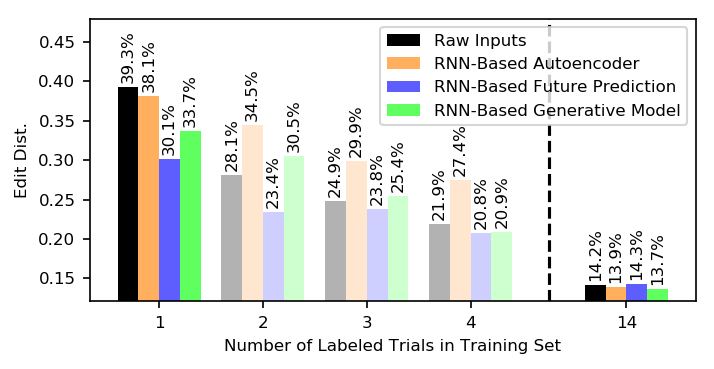}
\caption{\textbf{MISTIC-SL Maneuver Recognition: Edit distance vs. number of labeled trials.} The bottom of the $y$ axis is set to 12.1\%, the best published result using LSTM ($\sim$36 labeled trials). The non-transparent results are reported over exhaustive, deterministic splits (see Section \ref{experimentaldesign}) which can be compared to in future work.}
\label{misticeditbarplot}
\end{figure}

Here we study recognition performance in the annotation-scarce setting across two datasets and four approaches: recognition from raw inputs and recognition from learned representations using the autoencoder, the future-prediction model, and the full generative model. For recognition, we use the state-of-the-art model from \cite{dipietro2019}. We also carry over the same hyperparameters, which were optimized in \cite{dipietro2019} for recognition using raw inputs. Thus any improvements from using learned representations are not due to tuning.

\subsection{Datasets}
\label{datasets}

The two datasets used are the Minimally Invasive Surgical Training and Innovation Center -- Science of Learning dataset (MISTIC-SL) \cite{gao2016icra,gao2016ijcars} and the JHU-ISI Gesture and Skill Assessment Working Set (JIGSAWS) \cite{gao2014, ahmidi2017}. Both datasets consist of two distinct components: 1. measurements recorded automatically over time as a trainee operates the \emph{da Vinci}, including motion data, and 2. dense activity labels over time that were provided manually by experts.

MISTIC-SL focuses on recognizing maneuvers, with each activity label being 1 of 4 different manuevers: \emph{suture throw} (ST), \emph{knot tying} (KT), \emph{grasp pull run suture} (GPRS), or \emph{intermaneuver segment} (IMS). As input, we use 14 kinematic signals: velocities along the 3 axes, angular velocities along the 3 axes, and gripper angle, all for both the left and right hands. All signals are provided at 50Hz, which following \cite{dipietro2016,dipietro2019} we downsample by a factor of 6. The data was collected in a benchtop training environment. We follow \cite{gao2016icra,dipietro2016} and use 39 right-handed trials from 15 subjects, most of whom were residents.

JIGSAWS focuses on recognizing gestures -- short, low-level activities such as \emph{pushing needle through tissue} (see \cite{ahmidi2017} for a full list). We follow the majority of prior work and focus on the \emph{Suturing} task. Here each activity label is 1 of 10 different gestures. As input, we use the same 14 kinematic signals. Here, all signals are provided at 30Hz, and following \cite{dipietro2016,dipietro2019} we downsample by a factor of 6. The data was collected in a benchtop training environment and consists of 39 trials from 8 different subjects, most of whom were medical students.

\subsection{Experimental Design}
\label{experimentaldesign}

\begin{figure}[t]
\centering
\includegraphics[width=3.5in]{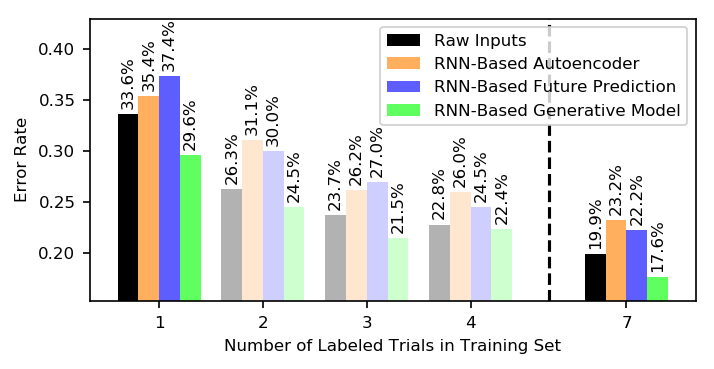}
\caption{\textbf{JIGSAWS Gesture Recognition: Error rate vs. number of labeled trials.} The bottom of the $y$ axis is set to 15.3\%, the best published result using LSTM ($\sim$35 labeled trials). The non-transparent results are reported over exhaustive, deterministic splits (see Section \ref{experimentaldesign}) which can be compared to in future work.}
\label{jigsawserrorbarplot}
\end{figure}

\begin{figure}[t]
\centering
\includegraphics[width=3.5in]{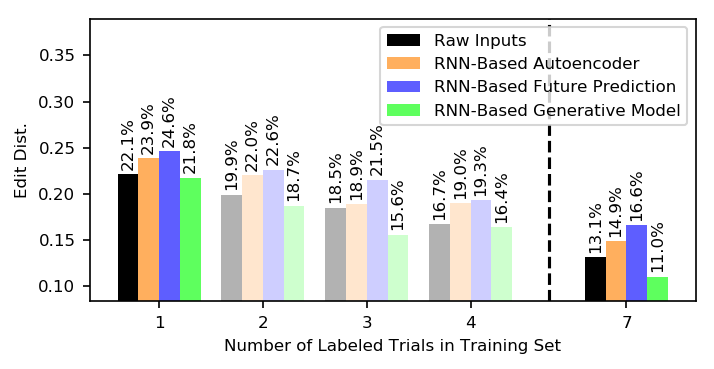}
\caption{\textbf{JIGSAWS Gesture Recognition: Edit distance vs. number of labeled trials.} The bottom of the $y$ axis is set to 8.4\%, the best published result using LSTM ($\sim$35 labeled trials). The non-transparent results are reported over exhaustive, deterministic splits (see Section \ref{experimentaldesign}) which can be compared to in future work.}
\label{jigsawseditbarplot}
\end{figure}

The recognition model from \cite{dipietro2019} consists of bidirectional LSTM with 3 layers, each with 64 hidden units, and is optimized using Adam and a learning rate of $10^{-2.5}$. All hyperparameters are carried over unchanged except the batch size: we use a batch size of 1 because it is the only possible option in many of our experiments. Training is carried out for 100 epochs. The metrics considered are frame-wise error rate and segment-wise edit distance (Levenshtein distance), normalized by the maximum number of segments in any one trial to aid interpretability, following prior work. In focusing on generalization across users, \emph{any particular training set consists of exactly one labeled trial per user}, with between 1 and $u - 1$ labeled trials, where $u$ is the number of users in the dataset. In all cases, results are averaged over splits, exhaustively for 1 trial and $u - 1$ trials and randomly otherwise (in this case 10 splits are randomly sampled).

During the representation-learning phase, the most important hyperparameters are the number of LSTM hidden units ($n_h$) and the number of components in the Gaussian mixture model ($n_c$). For the autoencoder and future-prediction models, these were tuned in \cite{dipietro2018} to maximize performance on a held-out set of 4 MISTIC-SL users, and here we use the same values ($n_h = 64, n_c = 16$). We followed the same process for the RNN-Based Generative Model. Tuning is carried out using the unsupervised-learning objective, not the recognition objective. This led to values of $n_h = 128, n_c = 8$. In all cases, training was carried out for 100 epochs using Adam, with a learning rate of 0.005.

\subsection{Results and Discussion}
\label{results}

Figure \ref{samples} shows examples of the gripper-angle signal from MISTIC-SL to illustrate the three unsupervised-learning tasks. Future prediction yields blurred, incoherent futures, whereas the generative model samples detailed trajectories. This suggests that the generative model's representations may be better suited for fine-grained activities; and this is confirmed below in the case of JIGSAWS.

Figure \ref{misticerrorbarplot} shows error rate vs. the number of labeled trials for maneuver recognition (MISTIC-SL). Using only one labeled trial, raw inputs for recognition lead to an error rate of 27.3\%; the autoencoder representations lead to an improvement, obtaining 23.9\%; and the future-prediction and generative models lead to further improvements, obtaining 20.3\% and 20.7\% respectively. Figure \ref{teaser} shows qualitative results using only 1 labeled trial with future prediction (results are similar for the generative model). When 14 labeled trials are used, the generative model yields the lowest error rate (10.1\%). For reference, the state-of-the-art LSTM result using 36 labeled trials is 8.7\% \cite{dipietro2019}. Figure \ref{misticeditbarplot} shows results for edit distance; the same general trends hold.

Figure \ref{jigsawserrorbarplot} shows error rate vs. the number of labeled trials for gesture recognition (JIGSAWS). Using only one labeled trial, raw inputs lead to an error rate of 33.6\%. Representations from the full generative model reduce the error rate to 29.6\%, while the autoencoder and future-prediction based representations both degrade performance. This is not surprising: we have no reason to believe that the autoencoder's task of signal reconstruction is well aligned with the task of activity recognition; and for future prediction, we have seen above that blurry, uncoherent futures are obtained, and this is likely detrimental to recognizing fine-grained activities such as gestures. When 7 labeled trials are used for training, the generative model again yields the lowest error rate (17.6\%). For reference, the state-of-the-art LSTM result using 35 labeled trials is 15.3\% \cite{dipietro2019}. Figure \ref{jigsawseditbarplot} shows results for edit distance; the same general trends hold.

\section{Conclusions and Future Work}
\label{conclusionsandfuturework}

Automated activity recognition in the presence of few annotations is an important problem which we believe warrants more attention. This work presented the first analysis of surgical activity recognition in this setting. We found that recognition is feasible using only one annotated sequence, and that leveraging learned representations, obtained in an unsupervised fashion, leads to performance boosts at recognition time. The RNN-based generative model introduced in this work is particularly strong for this purpose, especially when recognizing activities that occur over short time scales, as in gesture recognition. That said, a significant gap still exists in performances when fewer annotated sequences are available for training. We hope that the community will join us in seeing how much we can improve performance in this important annotation-limited regime.

\paragraph{\textbf{Acknowledgements}} This work was supported by a fellowship for modeling, simulation, and training from the Link Foundation. We also thank Anand Malpani, Madeleine Waldram, Swaroop Vedula, Gyusung I. Lee, and Mija R. Lee for procuring the MISTIC-SL dataset. The procurement of MISTIC-SL was supported by the Johns Hopkins Science of Learning Institute.

\bibliographystyle{splncs04}
\bibliography{paper1372}

\end{document}